\documentclass{article}

\usepackage[final]{AdvML_Frontiers_2024}
\usepackage{kotex}

\usepackage[utf8]{inputenc} % allow utf-8 input
\usepackage[T1]{fontenc}    % use 8-bit T1 fonts
\usepackage{hyperref}       % hyperlinks
\usepackage{url}            % simple URL typesetting
\usepackage{booktabs}       % professional-quality tables
\usepackage{amsfonts}       % blackboard math symbols
\usepackage{nicefrac}       % compact symbols for 1/2, etc.
\usepackage{microtype}      % microtypography
\usepackage{amsmath}
\usepackage{multirow}
\usepackage{graphicx}
\usepackage[toc,page]{appendix}
\usepackage{subcaption}
\usepackage{floatrow}
\usepackage{tablefootnote}
\usepackage[table,xcdraw]{xcolor}

\title{An Adversarial Learning Approach to \\ Irregular Time-Series Forecasting}

\author{%
  Heejeong Nam\thanks{First author and corresponding author} \\
  \texttt{hatbi2000@yonsei.ac.kr} \\
  \And
  Jihyun Kim \\
  \texttt{20160131@dongduk.ac.kr} \\
  \And
  Jimin Yeom \\
  \texttt{jyeom32@gatech.edu} \\
}

\begin{document}

%  완전 결과 extensive하게 가자. 이미 파일 다 뽑아뒀으니. 그리고 기존 work 그림이나 표로 정리하자. add on 하지말고 걍 젤 잘 나왔던 lstm + nr + sum estimator해보자. 

\maketitle

\begin{abstract}
Forecasting irregular time series presents significant challenges due to two key issues: the vulnerability of models to mean regression, driven by the noisy and complex nature of the data, and the limitations of traditional error-based evaluation metrics, which fail to capture meaningful patterns and penalize unrealistic forecasts. These problems result in forecasts which are often misaligned with human intuition. To tackle these challenges, we propose an adversarial learning framework with a deep analysis of adversarial components. Specifically, we emphasize the importance of balancing the modeling of global distribution \textit{(overall patterns)} and transition dynamics \textit{(localized temporal changes)} to better capture the nuances of irregular time series. Overall, this research provides practical insights for improving models and evaluation metrics, and pioneers the application of adversarial learning in the domain of irregular time-series forecasting.

\end{abstract}

\section{Introduction}

\begin{figure*}[tb!]
\centering{\includegraphics[width=\textwidth]{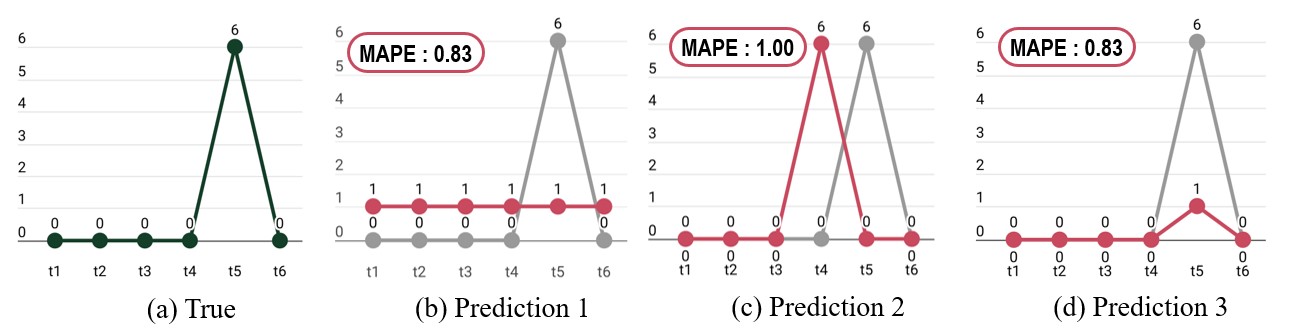}}
\caption{\textbf{Problems. } MAPE, where lower values indicate better performance, is highly unsuitable for evaluating the quality of irregular time-series forecasting. The true sequence (a) exhibits an irregular pattern. (b) matches the total quantity but reflects the monotonous tendencies common in many statistical models. (c), despite experiencing a slight temporal shift, receives the worst score. (d) provides a plausible result, capturing the correct entry point, yet it achieves the same MAPE as (b). \vspace{-10pt}}
\label{fig:issue}
\end{figure*}

Irregular time series, characterized by significant variations in inter-arrival times and quantities, pose unique challenges in analysis and forecasting. Unlike stationary and regular time series, which have seen substantial advancements in both methodologies and applications \cite{15_from_HCISS}, the exploration of irregular time series remains limited due to their inherent low interpretability. Forecasting such data is particularly challenging for two primary reasons—one stemming from the inadequacies of evaluation metrics and the other from the limitations of models, with both factors compounding each other. The first challenge lies in the widespread reliance on error-based metrics in forecasting, such as MAPE (mean absolute percentage error), which are ill-suited for capturing the unique characteristics of lumpy or intermittent patterns \cite{spec_martin, HCISS_compare}. These metrics fail to penalize the unrealistic forecasts often produced by statistical models, while simultaneously overlooking the potential of models that accurately identify underlying patterns but exhibit minor temporal shifts. The second challenge stems from the mean regression problem faced by forecasting models. This issue primarily arises due to the inherently noisy nature of irregular time series, which often lack clear trends or seasonality, making them especially prone to this problem. Paradoxically, under commonly used metrics like MAPE, the mean regression problem is not penalized but often rewarded, leading to models that fail to align with human intuition. Fig. \ref{fig:issue} provides a clear example where statistical models or models with mean regression problem deviate notably from intuitive expectations, despite achieving high scores with MAPE-based evaluations. Similar cases in real world datasets are detailed in Appendix \ref{appendix:real}. While fields such as large language models and image generation have made significant strides in producing outputs that align with human intuition, traditional forecasting models and evaluation metrics lag behind, especially in irregular time-series. To address this issue, we propose an adversarial learning framework aimed at bridging the gap between forecasting outputs and intuitive expectations. Adversarial learning, widely recognized for its applications in time-series generation and forecasting, is predominantly inspired by generative adversarial networks (GANs) \cite{gan}. While GANs have demonstrated strong capabilities in generating realistic and detailed outputs, their application to time-series forecasting remains underexplored particularly in the context of irregular data. In the domain of time-series forecasting, discriminators typically fall into one of two architectural paradigms: recursive layers, which align with conventional methods for sequential data processing \cite{c-rnn-gan, timegan, gtgan}, and non-recursive layers integrated with activation functions \cite{ast, traffic_cikm}. Despite their widespread use, a comprehensive analysis of the role and effectiveness of these architectures in time-series forecasting remains limited. To address this, we reexamine adversarial components and their impact on forecasting performance for irregular time-series data. To further tackle the limitations of traditional evaluation metrics which often neglect forecast plausibility, we propose novel qualitative approaches for analyzing irregular time-series forecasting. Our findings indicate that the architectural design of the adversarial components should align with the characteristics of the time series, balancing the capture of \textbf{global style} and \textbf{transition dynamics}. Finally, our research provides actionable insights into improving the models and the evaluation metrics for irregular time-series data. To our best knowledge, this is the first use of adversarial approaches in irregular time-series forecasting. We applied our approach to three real-world datasets, and our implementation is available at \href{https://github.com/Hazel-Heejeong-Nam/adversarial-intermittent-lumpy-forecasting}{Hazel-Heejeong-Nam/adversarial-intermittent-lumpy-forecasting}

\vspace{-5pt}

\section{Related Works}
\vspace{-5pt}
\paragraph{Adversarial learning in time series forecasting.}
Adversarial learning is often adopted in time series generation and forecasting \cite{gtgan, ast, timegan, traffic_cikm, stockgan, c-rnn-gan, other_cikm}. However, despite its widespread use, a thorough analysis of the role of adversarial learning in the time-series domain remains lacking. The unidirectional nature of sequential data introduces unique challenges, particularly in forecasting rather than in generation. Hence, it is crucial to revisit the objective of using adversarial components in forecasting, given the differing nature of these tasks. The earlier works including C-RNN-GAN \cite{c-rnn-gan} simply replaced the generator and discriminator with LSTM networks. On the other hand, RCGAN \cite{rcgan} conditioned on additional input instead of previous outputs, while it still utilized recurrent units for both the generator and discriminator. TimeGAN \cite{timegan} employed both supervised and adversarial objectives to address the mean regression problem and the lack of temporal dynamic consideration, respectively. Alongside other works \cite{5_from_timegan, 6_from_timegan}, TimeGAN integrated recurrent networks into both the generator and discriminator. GT-GAN \cite{gtgan}, while not focused on forecasting, employs GRU-ODE \cite{gru-ode} to construct the discriminator for generative purposes. AST \cite{ast} is the first adversarial model designed for forecasting, and it incorporates adversarial learning to prevent error accumulation from the autoregressive nature of predictions. Thus, the discriminator module consists of non-recursive layers and competes with the forecaster, which is expected to eliminate error accumulation. TrendGCN \cite{traffic_cikm} employs a Graph Convolutional Network as a forecaster and utilized two discriminators, one for capturing spatial relations and the other for temporal relations.
\vspace{-5pt}
\paragraph{Irregular time series forecasting.}
Irregular and sparse time-series forecasting is challenging as it is often characterized by multiple inter-arrival times, with some additionally distinguished by significant variations in quantity between these intervals. Although statistical methods (e.g., Croston \cite{croston}), machine learning approaches (e.g., SVR \cite{HUA}), and deep learning techniques (e.g., LSTM \cite{lstm}) have been applied to these problems \cite{HCISS_compare}, it remains unclear which method is the most suitable. This ambiguity arises not only due to insufficient investigation \cite{15_from_HCISS} but also because of the lack of appropriate metrics \cite{spec_martin, HCISS_compare}. Statistical methods like Croston \cite{croston}, Holt-Winters \cite{holtwinters}, and ARIMA \cite{arima} have shown strong performance in M4 forecasting competitions \cite{m4}. Among these, Croston remains particularly effective for forecasting intermittent and lumpy data. In machine learning, SVR \cite{svr} has been identified as a strong performer; for example, Hua et al. \cite{HUA} combined SVR with logistic regression. In deep learning, LSTM \cite{lstm} models have shown promise, although their performance can vary. While neural networks may underperform compared to statistical methods on metrics like RMSE (root mean squared error) and MAPE, they may outperform in terms of service-level quality. For example, MAPE and RMSE have been criticized for their inability to account for shifts and temporal interactions \cite{HCISS_compare}, which are strengths of neural networks.% \vspace{-5pt}

\section{Rethinking Adversarial Components in Irregular Time-Series Forecasting} \label{rethink}

\vspace{-5pt}
\begin{figure*}[tb!]
\centering{\includegraphics[width=\textwidth]{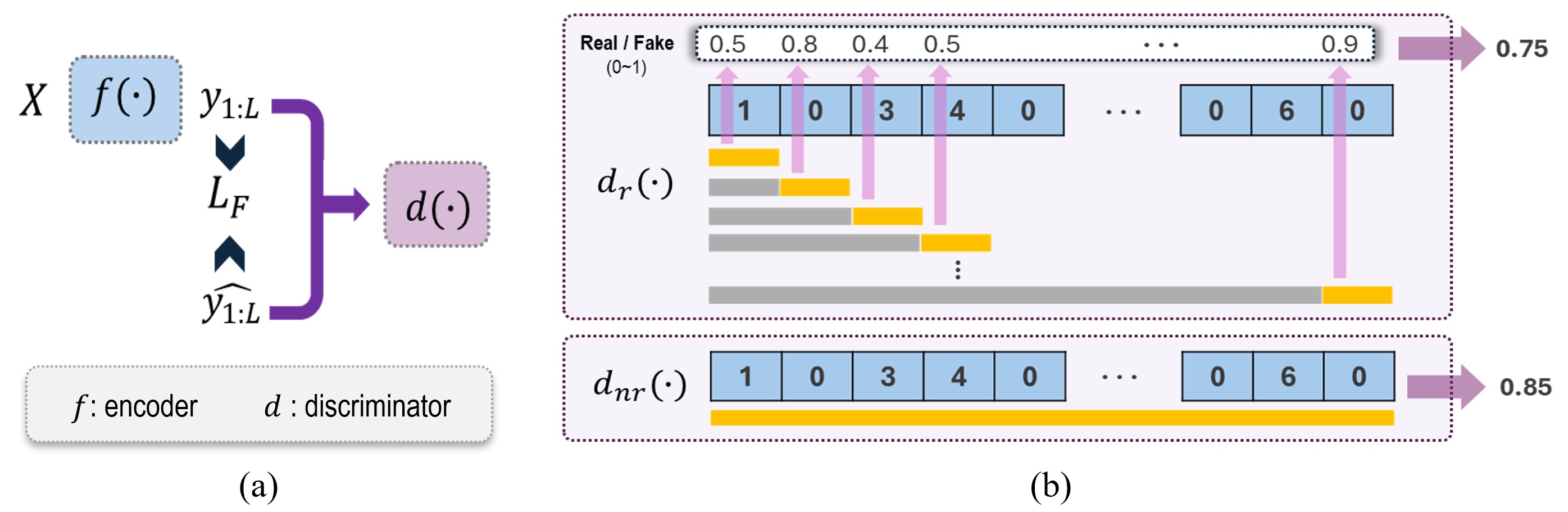}}
\caption{\textbf{Overall frameworks. } (a) Simple encoder-discriminator framework of adversarial learning. (b) Concept of $d_r(\cdot)$ (recursive discriminator) and $d_{nr}(\cdot)$ (non-recursive discriminator).  \vspace{-10pt}}
\label{fig:main}
\end{figure*}

\subsection{Problem Formulation}\label{pf} % 완료
\vspace{-5pt}
Let's assume we have \(\mathbf{M}\) distinct time series, each spanning a time period indexed by \(t = 1, \dots, T\). To forecast, models utilize historical data of length \(\mathbf{P}\), represented by \(\{x_{im}\}_{m=1}^P\) for each \(i^{\text{th}}\) time series. The forecasting horizon is set to \(\mathbf{L}\), and our objective is to accurately predict the values \(\{y_{in}\}_{n=1}^L\). Given a forecasting model \(\mathbf{F}:\mathbb{R}^P \rightarrow \mathbb{R}^L\), we express our forecasting process as \(\{\hat{y_{in}}\}_{n=1}^L=\mathbf{F}(\{x_{im}\}_{m=1}^P)\). In this paper, we only considered a global forecasting model across the different time series \(i=1, \dots, \mathbf{M}\) within the same dataset, i.e., the observational space is \(\mathcal{X} \in \mathbb{R}^{M \times T}\) while \(\mathbb{X} = \{x_{i,1:P}\}_{i=1}^M\) and \(\mathbb{Y} = \{y_{i,1:L}\}_{i=1}^M\) represent the sets of historical data and corresponding target values, respectively. Below, we specify the meaning of irregular time-series in our work.

\vspace{3pt}
\noindent
\textbf{Definition 3.1 \quad \textit{(Irregular Time-Series)}}\label{def} \quad \textit{We define irregular time series by their characteristic variability in inter-arrival times and further subcategorize them based on variability in quantity, drawing on the work of Syntetos et al. \cite{sbc}. Irregular time series are identified by using the \textbf{ADI} (Average Demand Interval), a metric that quantifies regularity over time by calculating the average interval between successive non-zero entries. Following established conventions \cite{HCISS_compare, sbc,  renewal}, we adopt a threshold of 1.32 for ADI ($ADI \geq 1.32$). Irregular time series can be further classified into two subcategories—\textbf{intermittent} and \textbf{lumpy}—based on the \(CV^2\) metric, which measures the variability in non-zero quantities. We use a threshold of 0.49 for \(CV^2\), as suggested in previous studies \cite{HCISS_compare, sbc,  renewal}.}
\vspace{-3pt}
\subsection{Adversarial Components}\label{sec:adver}
\vspace{-5pt}
\paragraph{Non-recursive Discriminator} Discriminators with non-recursive layers treat irregular time-series as vectors without considering temporal relationships, similar to the approach used in image transfer or generation \cite{stylegan, cyclegan}. For a fixed forecaster \(\boldsymbol{F}\), the non-recursive discriminator \(D_{nr}(\mathbf{y}, \theta_d)\) outputs a scalar which indicates the probability of input vector $\mathbf{y}$ originating from $\mathcal{X}$. We expect \(D_{nr}\) to operate as discribed in \eqref{eq:d_optimal}, and the global minimum is achieved if and only when \(p_F = p_{\mathcal{X}}\).
% \vspace{-8pt}
\begin{equation}
\label{eq:d_optimal}
D_{nr}(y_{1:L}) = \frac{p_{\mathcal{X}}(y_{1:L})}{p_{\mathcal{X}}(y_{1:L}) + p_F(y_{1:L})}
\end{equation}

\paragraph{Recurvise Discriminator} A discriminator with recursive layers $D_r$ functions similarly to a chain of multiple PCL modules \cite{pcl}. The PCL framework constructs two samples of vectors as shown in \eqref{eq:yandystar}. Here, \(v(n)\) intuitively provides a minimal description of the temporal dependencies in the data. For comparison, a fake data (in the context of adversarial learning) is sampled (\textit{forecasted} in our case), and the objective is to learn to discriminate between \(v(n)\) and \(v^*(n)\) using logistic regression as expressed in \eqref{eq:regression}, where \(h_n\) are scalar-valued functions that provide representation.
% \vspace{-4pt}
\begin{equation}
\label{eq:yandystar}
v(n) = \binom{y_{n-1}}{y_n} \text{\quad and \quad} v^*(n) = \binom{y_{n-1}}{y_n^*}
\end{equation}
\vspace{-4pt}
\begin{equation}
\label{eq:regression}
D_r(v) = \frac{1}{L}\sum_{n=1}^{L}B_n(h_n(v^1),h_n(v^2))
\end{equation}
\vspace{-4pt}

It has been theoretically proven \cite{pcl} that if the underlying sources of the time-series are 1) mutually independent, 2) temporally dependent, and 3) stationary, then \(h_n(y_{n-1})\) can recover the underlying sources up to permutation and component-wise transformation. However, in our Def.\ref{def}, it is challenging to assume stationary sources : for example, one might intuitively model one of the nonstationary sources as active and inactive phases, as proposed by Song et al .\cite{nctrl}, given that our dataset is characterized by varying inter-arrival times. Nevertheless, incorporating $D_r$ could significantly enhance the model's ability to capture and adapt to the underlying transition dynamics, providing a more robust framework for modeling the temporal shifts in the data.

% \vspace{-10pt}
\section{An adversarial approach to irreuglar time-series forecasting} \label{exp:additional}

\subsection{Dataset}
\begin{figure}\BottomFloatBoxes
\floatsetup{captionskip=4pt}
\begin{floatrow}
\ttabbox[]
  {\centering
   \tabcolsep=0.04cm
    \renewcommand{\arraystretch}{1.4}
    \small
%=========================================
       \begin{tabular}{cccc}
    \hline
                                   & \textbf{AUTO} & \textbf{M5} 
& \textbf{RAF} \\ \hline \hline
    Number of Time-Series (\(\mathbf{M}\)) & 1227          & 30489       
& 5000         \\ \hline
    Time Series Length (\(\mathbf{T}\))    & 24            & 182        
& 84           \\ \hline
    Mean Demand Interval  & 1.29         & 4.212           
& 10.02       \\ \hline
    $\text{CV}^2$& 4.38& 2.91
& 10.67\\ \hline
    Lumpy  & 286       & 5964        
& 2403         \\ \hline
    Intermittent  & 941     & 23040       & 2597         \\ \hline
    \end{tabular}
%=========================================
    }%
    {\caption{Summary of the key statistics for each dataset. For calculating $\text{CV}^2$ of demand size, the number of lumpy time-series and intermittent time-series, we follow SBC \cite{sbc}}
   \label{tab:stat}}%

    \killfloatstyle\ffigbox[]{
        \includegraphics[width=\linewidth]{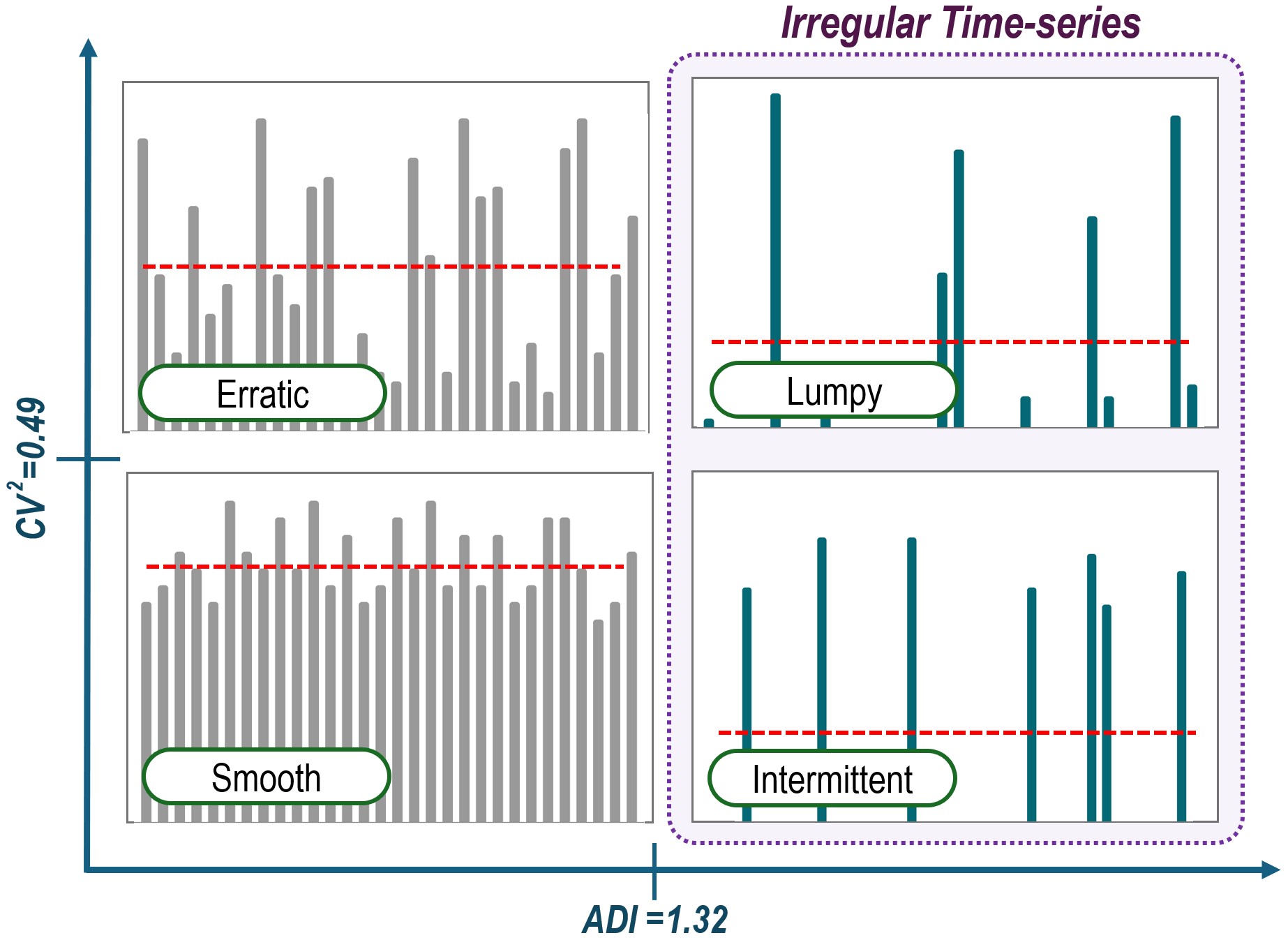}\vspace{-10pt}
    }
    {
        \caption{Illustration of irregular time-series}
        \label{fig:oursA}
        \vspace{-2pt}
    }
\end{floatrow}
\end{figure}

\vspace{-5pt}
All experiments in this paper were evaluated on three real-world datasets. Following Def.\ref{def}, we only include time-series classified as irregular according to the Syntetos-Boylan Classification \cite{sbc}. \textbf{AUTO} dataset includes monthly demand data for 3,000 items over 24 months. Among these, we are able to get 1,227 irregular time-series. \textbf{RAF} dataset comprises aerospace parts demand data from the Royal Air Force and it covers 84 months of data for 5,000 parts. Lastly, \textbf{M5} dataset includes daily unit sales per product and store at Walmart over 5 years and we used last 182 timestamps for our experiment. When comparing these datasets, the RAF dataset has the highest ADI and \(CV^2\) values, making it the most challenging for traditional forecasting methods. The AUTO dataset, while having fewer zero entries, exhibits greater variation in demand size. In contrast, the M5 dataset has less demand size variability but a higher number of zero entries compared to the AUTO dataset. Further details about these datasets are provided in Table \ref{tab:stat}.

% \vspace{-7pt}

\subsection{Evaluation}
\vspace{-5pt}
\paragraph{Conventional evaluation metrics.}
We first employed four metrics that are widely used. MAPE (Mean Absolute Percentage Error) is commonly used metric although there is ongoing debate \cite{HCISS_compare, spec_martin, traffic_cikm} regarding the suitability, as discussed in Fig. \ref{fig:issue}. Since MAPE has a fundamental limitation due to its asymmetry between predictions and true values, we also adopted sMAPE (Symmetric Mean Absolute Percentage Error) to mitigate this issue. Another widely used metric, RMSE (Root Mean Square Error), can effectively compare the similarity between vectors or matrices. However, it lacks the ability to account for temporal shifts often observed in time-series data. Furthermore, we include SPEC (Stock-keeping-oriented Prediction Error Cost) \cite{spec_martin}, a metric designed for demand forecasting which incorporates penalties for overstock and understock situations balanced by parameters $\alpha_1$ and $\alpha_2$ (we set both to 0.5). All equations are provided in Appendix \ref{appendix:eval}.

\vspace{-7pt}
\paragraph{Evaluation beyond error.} To address the limitations of existing metrics, we adopted several alternative approaches. The first metric, MSTD (Mean Standard Deviation), measures the alignment between the forecasts \(\hat{y_{i,1:L}}\) and the actual values \(y_{i,1:L}\). This metric evaluates how well the forecasting results capture the marginal distribution of the training dataset. Achieving a desirable MSTD indicates robustness against minor shifts in values (e.g., Fig. \ref{fig:issue} (c)) and focuses on reflecting the inherent patterns and statistical properties of the dataset, thus mitigating the mean regression problem. The second approach evaluates the ability to realistically represent void spaces, specifically the intervals in irregular time series where no entries exist. To assess how accurately these intervals are forecasted, we used the recall (V-Recall) and the F1 score (V-F1). A high V-Recall indicates that the model successfully predicts the absence of values in areas where no data is expected, while a strong V-F1 score demonstrates this capability while minimizing errors in interval predictions.

\vspace{-5pt}
\subsection{Adversarial Components in Irregular Time-Series Forecasting} \label{exp:found}
\vspace{-5pt}

\begin{table}[t!]
\ttabbox[]
  {\centering
   \tabcolsep=0.1cm
    \renewcommand{\arraystretch}{1.1}
    \small

% \begin{tabular}{@{}c|cccc|cccc|cccc@{}}
% \toprule
% \textbf{} & \multicolumn{4}{c|}{\textbf{AUTO}} & \multicolumn{4}{c|}{\textbf{RAF}} & \multicolumn{4}{c}{\textbf{M5}}  \\ \midrule
% \textbf{$F$} &
%   \multicolumn{2}{c}{MLP} &
%   \multicolumn{2}{c|}{LSTM} &
%   \multicolumn{2}{c}{MLP} &
%   \multicolumn{2}{c|}{LSTM} &
%   \multicolumn{2}{c}{MLP} &
%   \multicolumn{2}{c}{LSTM} \\ \midrule
% \textbf{$D$}     & MLP     & LSTM   & MLP    & LSTM   & MLP    & LSTM   & MLP    & LSTM   & MLP    & LSTM   & MLP   & LSTM   \\ \midrule
% MSTD    & 2.403   & 2.464  & 1.897  & 2.356  & 2.955  & 3.487  & 2.79   & 5.75   & 0.743  & 1.053  & 0.671 & 0.512  \\
% RMSE    & 7.822   & 9.728  & 7.922  & 7.608  & 16.855 & 20.399 & 16.609 & 27.485 & 2.693  & 2.575  & 2.317 & 3.145  \\
% MAPE    & 0.478   & 0.687  & 0.443  & 0.524  & 0.944  & 0.978  & 0.953  & 0.722  & 0.827  & 0.855  & 0.814 & 0.799  \\
% sMAPE   & 1.139   & 1.315  & 1.11   & 1.165  & 1.968  & 1.963  & 1.966  & 1.914  & 1.61   & 1.677  & 1.568 & 1.539  \\
% \textbf{SPEC}    & 2.005   & 2.05   & 1.517  & 1.926  & 1.534  & 0.348  & 0.803  & 0.72   & 24.144 & 30.644 & 18.25 & 26.949 \\ \bottomrule
% \end{tabular}

\begin{tabular}{@{}ccccccccccccc@{}}
\toprule
\textbf{}                          & \multicolumn{4}{c}{\textbf{AUTO}}                                     & \multicolumn{4}{c}{\textbf{RAF}}                                      & \multicolumn{4}{c}{\textbf{M5}}                    \\ \midrule
\multicolumn{1}{c|}{\textbf{$F$}}  & \multicolumn{2}{c}{MLP} & \multicolumn{2}{c|}{LSTM}                   & \multicolumn{2}{c}{MLP} & \multicolumn{2}{c|}{LSTM}                   & \multicolumn{2}{c}{MLP} & \multicolumn{2}{c}{LSTM} \\ \midrule
\multicolumn{1}{c|}{\textbf{$D$}}  & MLP    & LSTM           & MLP            & \multicolumn{1}{c|}{LSTM}  & MLP    & LSTM           & MLP           & \multicolumn{1}{c|}{LSTM}   & MLP    & LSTM           & MLP    & LSTM            \\ \midrule
\multicolumn{1}{c|}{MSTD}          & 2.403  & 2.464          & \textbf{1.897} & \multicolumn{1}{c|}{2.356} & 2.955  & 3.487          & \textbf{2.790} & \multicolumn{1}{c|}{5.75}   & 0.743  & 1.053          & 0.671  & \textbf{0.512}  \\
\multicolumn{1}{c|}{V-Recall} & 0.350  & \textbf{1.000} & 0.454          & \multicolumn{1}{c|}{0.350} & 0.874  & \textbf{0.972} & 0.902         & \multicolumn{1}{c|}{0.692}  & 0.861  & \textbf{0.896} & 0.799  & 0.661           \\
\multicolumn{1}{c|}{V-F1}     & 0.373  & \textbf{0.551} & 0.419          & \multicolumn{1}{c|}{0.373} & 0.896  & \textbf{0.945} & 0.911         & \multicolumn{1}{c|}{0.788}  & 0.742  & \textbf{0.764} & 0.728  & 0.672           \\ \midrule
\multicolumn{1}{c|}{RMSE}          & 7.822  & 9.728          & 7.922          & \multicolumn{1}{c|}{7.608} & 16.855 & 20.399         & 16.609        & \multicolumn{1}{c|}{27.485} & 2.693  & 2.575          & 2.317  & 3.145           \\
\multicolumn{1}{c|}{MAPE}          & 0.478  & 0.687          & 0.443          & \multicolumn{1}{c|}{0.524} & 0.944  & 0.978          & 0.953         & \multicolumn{1}{c|}{0.722}  & 0.827  & 0.855          & 0.814  & 0.799           \\
\multicolumn{1}{c|}{sMAPE}         & 1.139  & 1.315          & 1.11           & \multicolumn{1}{c|}{1.165} & 1.968  & 1.963          & 1.966         & \multicolumn{1}{c|}{1.914}  & 1.61   & 1.677          & 1.568  & 1.539           \\
\multicolumn{1}{c|}{SPEC}          & 2.005  & 2.05           & 1.517          & \multicolumn{1}{c|}{1.926} & 1.534  & 0.348          & 0.803         & \multicolumn{1}{c|}{0.72}   & 24.144 & 30.644         & 18.25  & 26.949          \\ \bottomrule
\end{tabular}
  
  }%
    {\caption{The forecasting results based on layer changes in the encoder and discriminator. }\label{tab:rethink}\vspace{-5pt}}%
\end{table}%
We first conduct experiments to assess the roles of adversarial components, specifically the encoder and discriminator, each of which can be configured in one of two ways: recursive and non-recursive. The simple framework is illustrated in Fig. \ref{fig:main} (a). The encoder takes an input \(x_{1:P}\) and produces an output \(\hat{y_{1:L}}\). In the recursive configuration, the encoder is implemented using two layers of LSTM \cite{lstm} and fully-connected layer, while in the non-recursive one consists of four linear layers. Similarly, the discriminator is implemented using either two layers of LSTM in the recursive case or a four linear layers in the non-recursive case. For the non-recursive discriminator, a single logit is directly produced by the successive linear layers to determine whether the input is real or fake. In contrast, in the recursive discriminator, the LSTM generates a logit at each state which comes from each hidden states, then the average of these logits is computed after after passing through the final fully-connected layer. This approach can be interpreted as evaluating the realisticity of the transition at each step, given the history. The basic training method follows the approach used in GANs \cite{gan}. For details, please refer to the Appendix \ref{appendix:exp}.

\begin{figure*}[tb!]
\centering{\includegraphics[width=\textwidth]{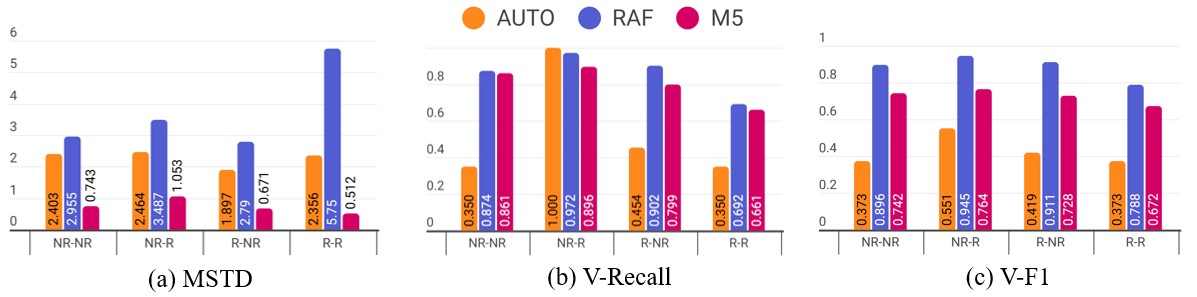}}
\caption{\vspace{-2pt}Visualization of three proposed metrics along different cofigurations. \vspace{-5pt}}
\label{fig:vis}
\end{figure*}

Across all datasets, the results in \ref{tab:rethink} emphasize the effectiveness of combining recursive and non-recursive architectures for irregular time series forecasting, enabling better modeling of both temporal dependencies and the marginal distribution. LSTM encoders paired with MLP discriminators tend to produce lower MSTD values, indicating better alignment with the marginal distribution of the data set. For V-Recall and V-F1, the opposite configurations consistently outperform, highlighting that LSTM discriminators excel at recognizing void intervals. Although the use of non-recursive MLP forecasters may seem unconventional in regular time series, they are particularly effective for irregular time series due to the absence of trends and seasonality. For conventional metrics, RMSE and MAPE generally favor configurations with LSTM encoders, although the effect of discriminator architecture appears less pronounced. Additionally, sMAPE and SPEC do not show consistent trends and appear to be more dataset-dependent.

% Detailed process of the framework can be found in Appendix \ref{appendix:exp}.
\vspace{-5pt}
\subsection{Comparison with Baseline Models}
\vspace{-5pt}

For the baseline models, we selected three statistical models (Croston \cite{croston}, ARIMA \cite{arima}, and ADIDA \cite{adida}) and three neural network models (MLP, RNN \cite{rnn}, and LSTM {lstm}). Each baseline models are either known for their strong performance on irregualr patterns or demonstrating simple architecture but showing strong performance. More detailed information about the baseline models can be found in the Appendix \ref{appendix:baseline}. The results in Table \ref{tab:main} show that, in terms of MSTD, statistical models generally underperformed, while simpler neural networks such as MLP, RNN, and LSTM exhibited moderate success. The best performance was achieved with adversarial learning, which closely matches the true variation in the dataset. Although the RAF dataset displayed an opposite trend, configurations of LSTM with non-recursive discriminator outperformed a pure LSTM by a significant margin in the other two datasets. Furthermore, V-Recall and V-F1 scores improved significantly with the adversarial approach, especially compared to other baseline models. This highlights the limitations of statistical models in irregular time series forecasting, emphasizing their unsuitability for capturing the complexities of such data. Table \ref{tab:main} presents results for the four primary metrics of interest, while the full results, including RMES, SPEC and sMAPE, are provided in Appendix \ref{fullresult}.

\begin{table}[t!]
\ttabbox[]
  {\centering
   \tabcolsep=0.06cm
    \renewcommand{\arraystretch}{1.1}
    \small

\begin{tabular}{@{}c|cccc|cccc|cccc@{}}
\toprule
\multirow{2}{*}{} & \multicolumn{4}{c|}{\textbf{AUTO}}                                & \multicolumn{4}{c|}{\textbf{RAF}}                                 & \multicolumn{4}{c}{\textbf{M5}}                                   \\ \cmidrule(l){2-13} 
                  & MSTD           & V-Recall       & V-F1           & MAPE           & MSTD           & V-Recall       & V-F1           & MAPE           & MSTD           & V-Recall       & V-F1           & MAPE           \\ \midrule
Croston \cite{croston}          & 2.403          & 0.027          & 0.049          & 0.507          & \textbf{2.653} & 0.520          & 0.663          & \textbf{0.844} & 1.270          & 0.580          & 0.661          & \textbf{0.637} \\
ARIMA \cite{arima}            & 2.294          & 0.228          & 0.285          & 0.557          & 2.661          & 0.748          & 0.827          & 0.926          & 1.198          & 0.845          & 0.766          & 0.786          \\
ADIDA \cite{adida}            & 2.403          & 0.039          & 0.069          & 0.477          & 2.653          & 0.706          & 0.798          & 0.919          & 1.270          & 0.646          & 0.702          & 0.676          \\ \midrule
MLP               & 2.273          & 0.421          & 0.411          & \textbf{0.404} & 2.664          & 0.845          & 0.881          & 0.982          & 1.072          & 0.895          & 0.766          & 0.844          \\
RNN \cite{rnn}              & 2.223          & 0.000          & 0.000          & 0.425          & 2.711          & 0.904          & 0.912          & 0.955          & 1.220          & 0.880          & 0.764          & 0.838          \\
LSTM \cite{lstm}             & 2.208          & 0.483          & 0.432          & 0.465          & 2.692          & 0.865          & 0.891          & 0.942          & 1.107          & 0.881          & \textbf{0.770} & 0.859          \\ \midrule
\textbf{R-NR}              & \textbf{1.897} & 0.454          & 0.419          & 0.443          & 2.79           & 0.902          & 0.911          & 0.953          & \textbf{0.671} & 0.799          & 0.728          & 0.814          \\
\textbf{NR-R}              & 2.464          & \textbf{1.000} & \textbf{0.551} & 0.687          & 3.487          & \textbf{0.972} & \textbf{0.945} & 0.978          & 1.053          & \textbf{0.896} & 0.764          & 0.855          \\ \bottomrule
\end{tabular}

      }%
    {\caption{Comparison results between the adversarial approach and other baselines.}\label{tab:main}}%
\end{table}%

\vspace{-5pt}

\section{Conclusion}
\vspace{-5pt}
% In conclusion, we have applied adversarial learning to forecast irregular time-series for the first time. We observed that well-performing existing methods often produce results that are overly unrealistic. Through a proof-of-concept experiment, we demonstrate that adversarial learning can address this issue, especially within irregular time-series, an area that has been underexplored in experimental analysis. While error-based metrics tend to favor conservative and safe forecasting, it might be worth considering for the importance of assessing forecasting quality in terms of capturing the unique style and marginal distribution of the dataset. Although increased variation in forecasting values can easily lead to a loss in error-based accuracy which is considered as limitation of our work, we believe that it hopefully paves the way for further development of metrics and methods in time-series forecasting that align more closely with human intuition.

In conclusion, this study represents the first application of adversarial learning for forecasting irregular time series. Our findings reveal that while existing methods often perform well according to conventional metrics, they frequently produce results that are overly simplistic or unrealistic. Through experiments, we demonstrate that adversarial learning can effectively address this issue, offering a more sophisticated approach for forecasting irregular time series. Although traditional error-based metrics tend to favor conservative forecasting models that prioritize accuracy, we argue that it is crucial to also assess forecasting quality through the lens of capturing the unique characteristics and marginal distribution of the data. The introduction of adversarial learning may lead to increased variation in forecasting values, which could result in a loss of accuracy in error-based metrics, yet we view this as an important step towards more flexible and realistic forecasting. This work, despite its limitations, lays the foundation for future research aimed at developing new metrics and methodologies that better capture the nuances of irregular time series data and more closely align with human intuition in forecasting tasks.

\newpage

%Bibliography
\bibliographystyle{unsrt}  
\bibliography{references}

%%%%%%%%%%%%%%%%%%%%%%%%%%%%%%%%%%%%%%%%%%%%%%%%%%%%%%%%%%%%
\newpage

\appendix

\section{Real-world Examples of Mean Regression Problem and MAPE in Irregular time Series}\label{appendix:real}

\begin{figure*}[h]
\centering{\includegraphics[width=\textwidth]{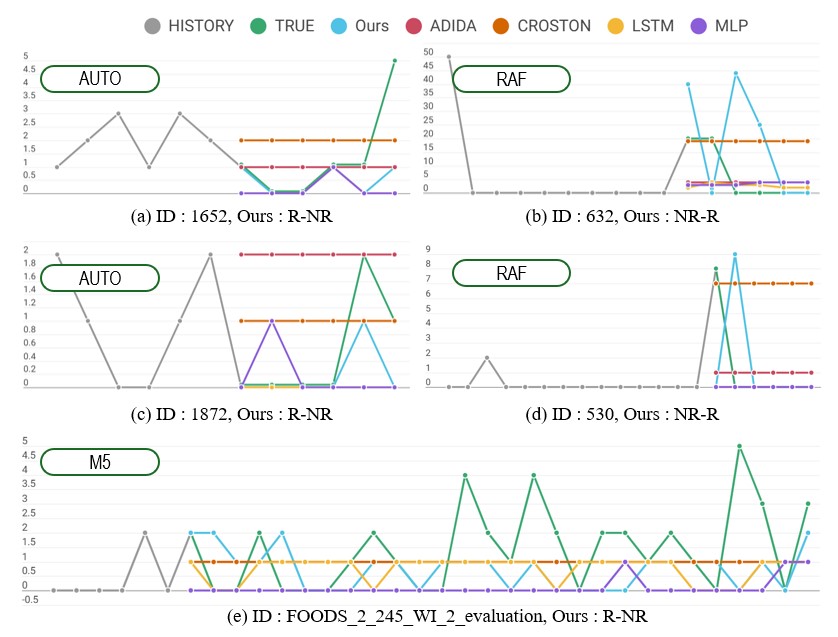}}
\caption{Real-world examples that demonstrate the existence of the problem we are addressing.}
\label{fig:realexamples}
\end{figure*}

\begin{table}[ht!]
\ttabbox[]
  {\centering
   \tabcolsep=0.2cm
    \renewcommand{\arraystretch}{1.0}
    \small

\begin{tabular}{@{}c|ccccc@{}}
\toprule[0.15em]
     & OURS & CROSTON & ADIDA & LSTM & MLP  \\ \midrule \midrule
(a)  & 0.45 & 0.2     & 0.9   & 0.75 & 0.75 \\ \midrule
(b)  & 1    & 0.8     & 0.05  & 0.85 & 0.85 \\ \midrule
(c)  & 0.75 & 0.75    & 0.25  & 1    & 1    \\ \midrule
(d)  & 1    & 0.875   & 0.125 & 1    & 1    \\ \midrule
(e)  & 0.06 & 0.05    & 0.01  & 0.06 & 0.06 \\ \bottomrule[0.15em]
\end{tabular}
  
  }%
    {\caption{Calculated MAPE in each example.}\label{tab:appreal}}%
\end{table}%

\section{Conventional Evaluation Metric} \label{appendix:eval}
\vspace{-5pt}

The following equations represent the methods used to evaluate our experimental results. The notation is consistent with that in Section \ref{pf}.

\begin{flalign}
    &MAPE = \frac{1}{LM}\sum_{i=1}^{M}\sum_{n=1}^{L}\frac{|y_{in}-\hat{y_{in}}|}{y_{in}})= \frac{1}{M}\sum_{i=1}^{M}(\frac{1}{\sum_{n=1}^{L}\mathbb{I}[y_{in}>0]}\sum_{n=1}^{L}\mathbb{I}[y_{in}>0]\frac{|y_{in}-\hat{y_{in}}|}{y_{in}})&
\end{flalign}
\vspace{-10pt}
\begin{flalign}
      &sMAPE = \frac{2}{LM}\sum_{i=1}^{M}\sum_{n=1}^{L}\frac{|y_{in}-\hat{y_{in}}|}{y_{in}+\hat{y_{in}}} \nonumber & \\
      &\quad \quad \quad \quad= \frac{2}{M}\sum_{i=1}^{M}(\frac{1}{\sum_{n=1}^{L}\mathbb{I}[y_{in}+\hat{y_{in}}>0]}\sum_{n=1}^{L}\mathbb{I}[y_{in}+\hat{y_{in}}>0]\frac{|y_{in}-\hat{y_{in}}|}{y_{in}+\hat{y_{in}}})  &
\end{flalign}
\vspace{-10pt}
\begin{flalign}
 &SPEC = \frac{1}{ML}\sum_{i=1}^{M}\sum_{n=1}^{L}\sum_{m=1}^{n}((n-m+1)\cdot max(0 ; \alpha_1 \cdot min(y_{im}; \sum_{k=1}^{m}y_{ik}- \sum_{j=1}^{n}f_{ij}); \nonumber & \\
 &\quad \quad \quad \quad\quad\alpha_2 \cdot min(f_{im}; \sum_{k=1}^{m}f_{ik}- \sum_{j=1}^{n}y_{ij}))) &
\end{flalign}
\vspace{-10pt}
\begin{flalign}
 &RMSE =  \frac{1}{M}\sum_{i=1}^M\sqrt{\sum_{n=1}^L\frac{(y_{in}-\hat{y_{in}})^2}{n}}&
\end{flalign}
\vspace{-10pt}
% \begin{flalign}
%  &STD MAE = \frac{1}{M}\sum_{i=1}^M |\text{stddev}(y_{i,1:L})-\text{stddev}(y_{i,1:L})|&
% \end{flalign}

\section{Baseline models}\label{appendix:baseline}

\begin{itemize}
\setlength\itemindent{-20pt} 
    \setlength\itemsep{5pt} % Optional: Adjust item spacing
    \item \textbf{Croston} \cite{croston}: A method specifically designed for intermittent and lumpy time series forecasting. It decomposes demand into occurrence and size, making it effective for irregular patterns in time-series data.
    
    \item \textbf{ARIMA} \cite{arima}: A widely-used statistical model that combines autoregressive (AR) and moving average (MA) components with differencing to handle non-stationarity. It is effective for capturing linear temporal dependencies in regular time series but struggles with highly irregular patterns.

    \item \textbf{ADIDA (Aggregate-Disaggregate Intermittent Demand Approach) \cite{adida}}: A technique tailored for intermittent time series. It aggregates demand over fixed intervals to smooth irregular patterns and then applies standard forecasting methods to generate predictions.

    \item \textbf{MLP (Multilayer Perceptron)}: A feedforward neural network model that learns non-linear relationships between input and output. While flexible, it may require careful tuning for effective performance on time series with irregular patterns.

    \item \textbf{RNN (Recurrent Neural Network) \cite{rnn}}: A neural network model with feedback connections that capture temporal dependencies in sequential data. RNNs are powerful for regular time series but can struggle with long-term dependencies or highly intermittent patterns.

    \item \textbf{LSTM (Long Short-Term Memory) \cite{lstm}}: A specialized type of RNN designed to overcome the vanishing gradient problem, making it suitable for learning long-term dependencies in sequential data. LSTM can handle some degree of irregularity but requires significant computational resources.
\end{itemize}

\section{Experiment details} \label{appendix:exp}

We first set look-back period and forecasting horizon for each dataset. We follow setting of kaggle competition for M5 dataset, thus both $P$ and $L$ are set to 28. AUTO and RAF are having $L$ of 6, while $P$ is set to 6 and 18 respectively. AUTO is having shorter look-back period due to the lack of historical data, as we only have 24 timestamp through out whole dataset.

In our experiments, we train encoder and discriminator jointly. Throughout all configurations, we trained for 100 epochs and hyperparameter tuning has been done in logarithmic scale. We selected our best model by using MAPE which can be considered as convention, and we observed how the models behave throughout in other metrics.  All seeds in the experiment is set to 0 and we set a batch size to 256. During training, we first update parameters in encoder followed by discriminator. 

\section{Additional Experiment Results} \label{fullresult}

\begin{table}[t]
\ttabbox[]
  {\centering
   \tabcolsep=0.06cm
    \renewcommand{\arraystretch}{1.3}
    \small

\begin{tabular}{@{}c|cccc|cccc|cccc@{}}
\toprule
\multirow{2}{*}{} & \multicolumn{4}{c|}{\textbf{AUTO}}     & \multicolumn{4}{c|}{\textbf{RAF}} & \multicolumn{4}{c}{\textbf{M5}} \\ \cmidrule(l){2-13} 
                  & MAPE  & RMSE  & sMAPE & SPEC           & MAPE   & RMSE    & sMAPE  & SPEC  & MAPE   & RMSE  & sMAPE & SPEC   \\ \midrule
Croston \cite{croston}           & 0.507 & 8.790 & 1.109 & 2.831          & 0.844  & 24.672  & 1.949  & 6.562 & 0.637  & 2.288 & 1.425 & 18.796 \\
ARIMA \cite{arima}             & 0.557 & 7.543 & 1.288 & 1.872          & 0.926  & 16.193  & 1.967  & 1.915 & 0.786  & 2.204 & 1.551 & 0.718  \\
ADIDA \cite{adida}            & 0.477 & 7.838 & 1.122 & 2.074          & 0.919  & 16.483  & 1.966  & 2.172 & 0.676  & 2.268 & 1.445 & 17.375 \\ \midrule
MLP               & 0.404 & 7.879 & 1.068 & 1.802          & 0.982  & 16.284  & 1.986  & 0.906 & 0.844  & 2.421 & 1.664 & 29.477 \\
RNN \cite{rnn}              & 0.425 & 7.911 & 1.089 & 1.632          & 0.955  & 16.623  & 1.974  & 1.189 & 0.838  & 2.253 & 1.647 & 25.398 \\
LSTM \cite{lstm}             & 0.465 & 8.020 & 1.129 & 1.904          & 0.942  & 16.649  & 1.973  & 1.680 & 0.859  & 2.422 & 1.691 & 31.623 \\ \midrule
R-NR              & 0.443 & 7.922 & 1.110 & 1.517          & 0.953  & 16.609  & 1.966  & 0.803 & 0.814  & 2.317 & 1.568 & 18.250 \\
NR-R              & 0.687 & 9.728 & 1.315 & 2.050          & 0.978  & 20.399  & 1.963  & 0.348 & 0.855  & 2.575 & 1.677 & 30.644 \\ \bottomrule
\end{tabular}

  }%
    {\caption{Additional results.}}%
\end{table}%

\vfill

\end{document}